\documentclass[10pt,twocolumn,letterpaper]{article}

\usepackage{cvpr}
\usepackage{times}
\usepackage{epsfig}
\usepackage{graphicx}
\usepackage{amsmath}
\usepackage{amssymb}
\usepackage{stfloats}
\pdfoutput=1

\usepackage[ruled]{algorithm2e}
\usepackage{subfigure}
\usepackage{bm}
\usepackage{makecell}
\usepackage{multirow}
\usepackage{threeparttable}

\usepackage[breaklinks=true,bookmarks=false]{hyperref}

\cvprfinalcopy 

\def\cvprPaperID{7708} 

\ifcvprfinal\fi
\begin{document}

\title{SQE: a Self Quality Evaluation Metric for Parameters Optimization in Multi-Object Tracking}


\author{%
Yanru Huang$^1$ \quad Feiyu Zhu$^2$ \quad Zheni Zeng$^1$\thanks{This work was done during Zheni Zeng was an intern at Megvii Inc.} \quad Xi Qiu$^2$ \quad Yuan Shen$^1$\thanks{Corresponding author.} \quad Jianan Wu$^2$\\
{$^1$Tsinghua University} \quad       
{$^2$Megvii Inc.}\\
{\texttt{\small{\{huangyr18, zengzn16\}@mails.tsinghua.edu.cn}} \hspace{0.5cm}}
{\texttt{\small{shenyuan\_ee@tsinghua.edu.cn}}\hspace{0.5cm}}\\
{\texttt{\small{\{zhufeiyu, qiuxi, wjn\}@megvii.com}}}
}

\maketitle
\thispagestyle{empty}

\begin{abstract}
We present a novel self quality evaluation metric ${\rm SQE}$ for parameters optimization in the challenging yet critical multi-object tracking task. Current evaluation metrics all require annotated ground truth, thus will fail in the test environment and realistic circumstances prohibiting further optimization after training. By contrast, our metric reflects the internal characteristics of trajectory hypotheses and measures tracking performance without ground truth. We demonstrate that trajectories with different qualities exhibit different single or multiple peaks over feature distance distribution, inspiring us to design a simple yet effective method to assess the quality of trajectories using a two-class Gaussian mixture model. Experiments mainly on MOT16 Challenge data sets verify the effectiveness of our method in both correlating with existing metrics and enabling parameters self-optimization to achieve better performance. We believe that our conclusions and method are inspiring for future multi-object tracking in practice.
\end{abstract}

\section{Introduction}

Multi-object tracking (MOT) aims to track all objects of interest categories in a video sequence \cite{schulter2017deep, zhang2019frame}.  It is crucial in applications like video surveillance and autonomous driving, where multiple pedestrians and vehicles need to be tracked simultaneously \cite{feng2019multi, yoon2019online, bernardin2008evaluating}. In recent years, tracking-by-detection \cite{lenz2015followme, schulter2017deep, wojke2017simple, feng2019multi, andriyenko2012discrete, breitenstein2010online} has become the predominant paradigm of MOT. This approach first detects objects in each frame, then extracts discriminative features to quantify the similarities between targets, and finally perform data association to assign detections into their most likely trajectories. During this process, several influential parameters need to be set manually, such as the threshold determining whether to establish associations. To find the optimal parameters, an evaluation procedure is needed to measure the tracking performance. However, existing evaluation metrics like event-based measures CLEAR MOT \cite{bernardin2008evaluating} or identity-based measures ${\rm IDF_1}$ \cite{ristani2016performance} all require ground truth annotations, limiting the optimization to training data. Since the optimized parameters could be sub-optimal in test scenes, a self evaluation metric that enables parameters optimization without ground truth is urgently needed.

\begin{figure}[t]
\centering
\includegraphics[width=1\linewidth]{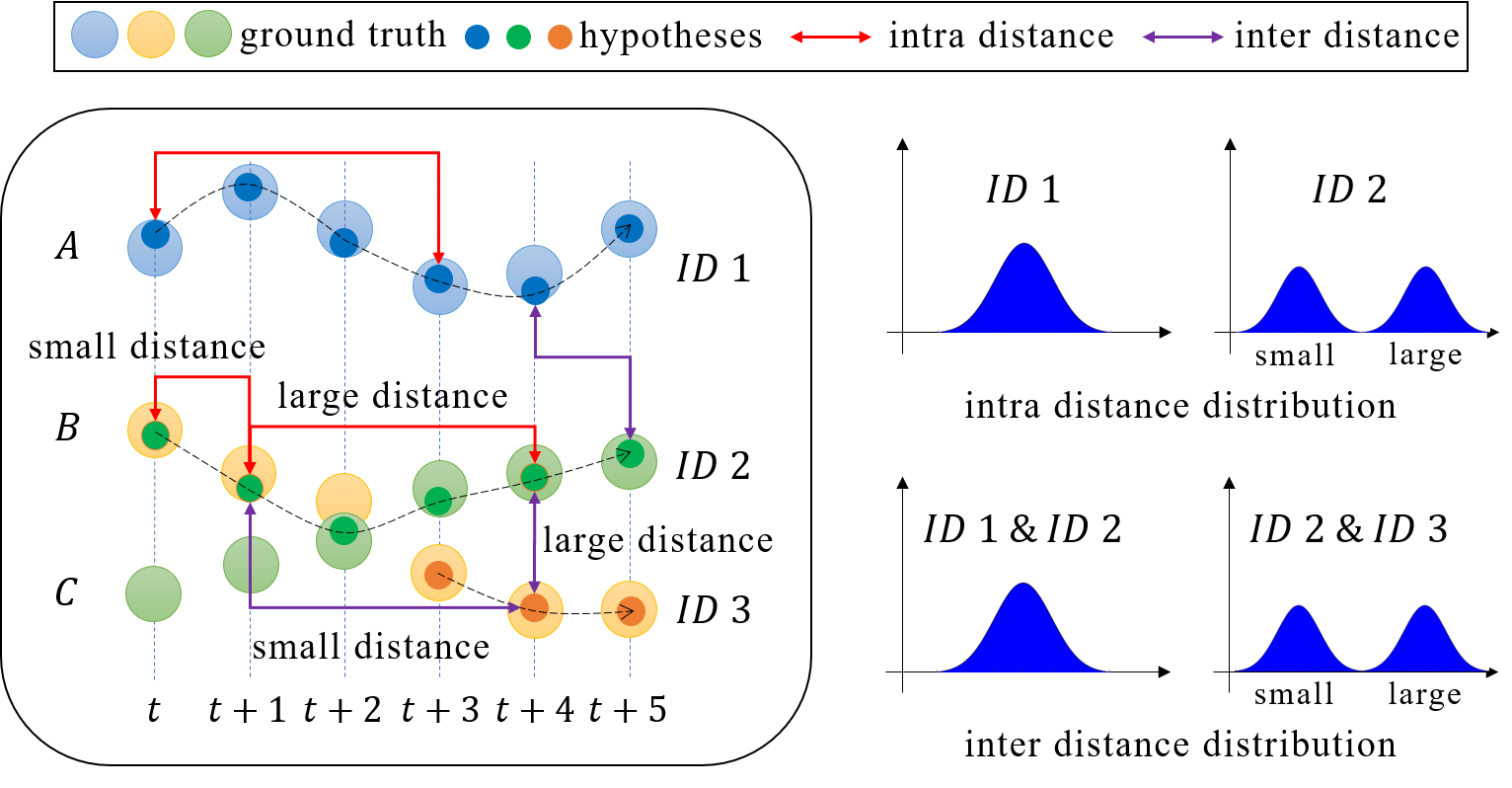}
\centering
\caption{Examples of different intra and inter distance distributions. Ideally, the distance distribution within the same trajectory or between two different trajectories present a single peak with little noise. If false identification takes place, such as the hypothetical identity 2 follows target B at first and then switches to another target C, the distance distribution would present multiple distinguishable peaks. See Section 3.2 for a detailed explanation.}
\label{fig1}
\end{figure}

To evaluate the accuracy and stability of a tracker without ground truth, we design a self quality evaluation metric ${\rm SQE}$ that considers the quantity, length, and feature distance information of the trajectory hypotheses comprehensively. Our method can assess the quality of trajectories owing to the distinctive distance distribution forms as shown in Figure \ref{fig1}. The intra distance denotes the feature distance between each two detection boxes in the same trajectory, and all pairs constitute the intra distance distribution. Similarly, the inter distance denotes the feature distance between each two detection boxes from different trajectories. Intuitively, when a trajectory contains different targets, the distance distribution scatters, and we demonstrate that it has the general characteristic of multiple peaks.

${\rm SQE}$ enables automatic parameters adaptation to accommodate different scenes. Designing a tracking algorithm that performs well under various video scenes is hard, yet tuning parameters in existing tracking algorithms can achieve equally outstanding performance in an easier manner. To the best of our knowledge, there is no previous work in this area to date. We believe that our approach is exceedingly instructive and provides new ideas for future research.

In summary, our contributions are as follows: (1) We show that feature distance distributions can reflect trajectory hypotheses quality; (2) We propose a self quality evaluation metric ${\rm SQE}$ based on two-class Gaussian mixture model, which can primarily fulfill the self-evaluation desire; (3) We test the effectiveness of our method on various data sets and note its drawbacks. A future prospect of using distributions to estimate erroneous frames is discussed in the end.

\section{Related Work}

\subsection{MOT algorithms}

In the tracking-by-detection paradigm, trackers first detect objects in each frame and then associate detections over time to form trajectories for targeted individuals \cite{lenz2015followme, zhang2019frame, milan2017online}. Online methods \cite{oh2004markov, xiang2015learning, bewley2016simple, wojke2017simple} only use previous and current frames and are thus suitable for real-time applications. One straightforward implementation is simple online and real-time tracking (SORT) \cite{bewley2016simple}, which predicts the new locations of bounding boxes using Kalman filter, followed by a data association procedure using intersection-over-union (IOU) to calculate the cost matrix. Although SORT achieves favourable speed and accuracy simultaneously, it suffers from heavy identity switches due to short-term motion information. Deep SORT \cite{wojke2017simple}, on the other hand, introduces object re-identification (REID) as appearance information to handle long-term occlusions, leading to a more robust and effective algorithm. Due to the rapid development of deep neural networks (DNNs), REID features with powerful discriminative capability have been popularized in MOT algorithms \cite{feng2019multi, tang2017multiple, zhang2019frame, yoon2019online}. In addition, the frame-by-frame association problem is often seen as bipartite graph matching solved by Hungarian algorithm \cite{kuhn1955hungarian}.

By contrast, offline methods \cite{zhang2008global, lenz2015followme, berclaz2011multiple} have access to the whole sequence and can perform global optimization on data association. These batch methods generally formulate MOT as a network flow problem \cite{zhang2008global, pirsiavash2011globally}. K-shortest paths (KSP) \cite{berclaz2011multiple}, successive shortest-path (SSP) \cite{lenz2015followme}, and dynamic programming (DP) \cite{pirsiavash2011globally} can be used to find the optimal solution. Offline methods enable correction of early errors in online methods and often show better performance, but are not applicable to time-critical applications.

In this paper, we focus on a simple, efficient, and easy-to-implement tracking framework. We use REID features to calculate the cost between current object detections and existing tracklets, minimize the total cost by Hungarian algorithm, and employ operations like interpolation and merging to correct previous results. Among all the parameters that need to be set, the REID threshold and merging threshold are the two most dominant parameters, which allows establishing associations and merging tracklets respectively.

\subsection{Evaluation metrics}

\begin{figure}[t]
\centering
\includegraphics[width=1\linewidth]{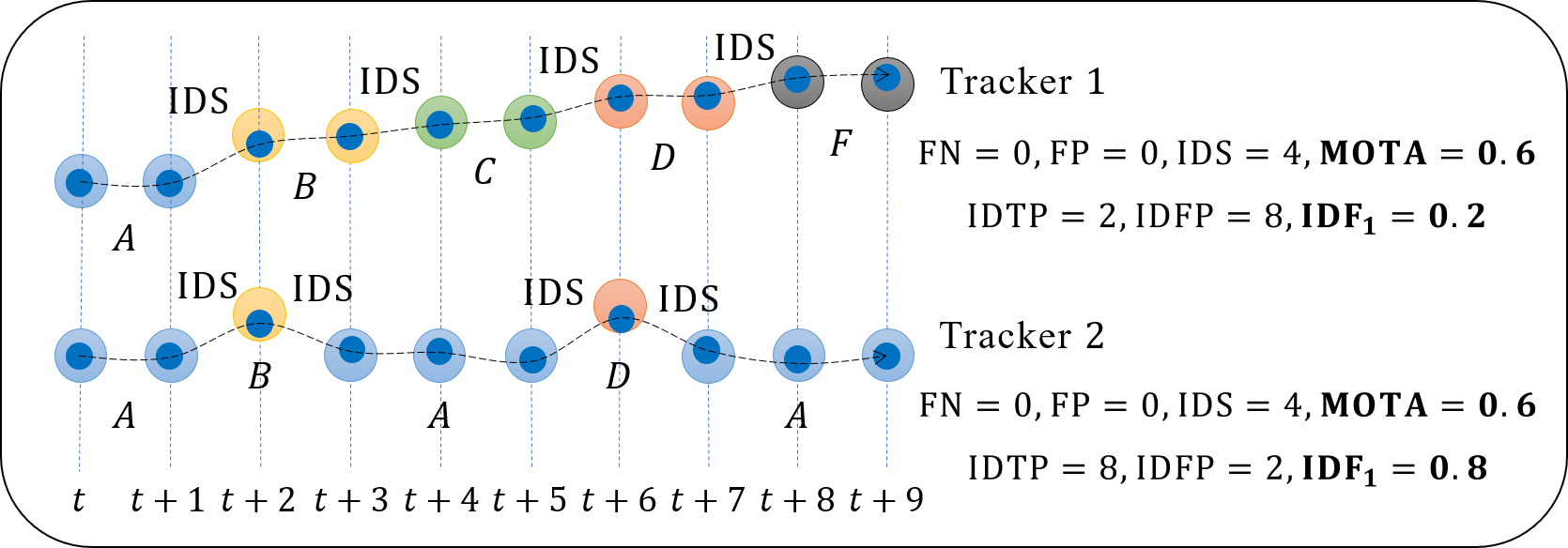}
\centering
\caption{Comparison of ${\rm MOTA}$ and ${\rm IDF_1}$. Both Tracker 1 and Tacker2 track target A which lasts 10 frames. Tracker 1, which correctly tracks 2 frames and the remaining frames are assigned to other wrong targets, has 4 identity switches, while Tracker 2 tracks 8 frames with 2 errors has the same number of identity switches. ${\rm IDF_1}$ measures how long the identification is correct and give different scores, which can better reflect trackers' performance.}
\label{fig2}
\end{figure}

Quantitative evaluation of tracking performance is challenging due to the complexity of multi-target tracking task. A large number of metrics have been proposed \cite{li2009learning, schuhmacher2008consistent, smith2005evaluating, nghiem2007etiseo}, including two main common metrics serving different purposes. One of them is CLEAR MOT metrics \cite{bernardin2008evaluating, milan2016mot16}, which contains multiple object tracking accuracy (MOTA) and multiple object tracking precision (MOTP):
\begin{equation}\label{eq1}
{\rm MOTA}=1-\dfrac{\sum_t({\rm FN}_t+{\rm FP}_t+{\rm IDS}_t)}{\sum_t{\rm GT}_t},
\end{equation}
\begin{equation}\label{eq2}
{\rm MOTP}=\dfrac{\sum_{i,t}d_t^i}{\sum_tc_t},
\end{equation}
where $c_t$ denotes the number of matched targets in frame $t$, and $d_t^i$ denotes the matching distance of target $i$. Comparing to {\rm MOTP}, which is mainly influenced by localization accuracy of detections, {\rm MOTA} sums various sources of errors, including false negatives, false positives, and identity switches, providing a better overall performance measure.

The other is ID metrics \cite{ristani2016performance}, which contains identification precision (${\rm IDP}$), identification recall (${\rm IDR}$) and corresponding ${\rm F_1}$ score ${\rm IDF_1}$:
\begin{equation}\label{eq3}
{\rm IDF_1}=\dfrac{2{\rm IDTP}}{2{\rm IDTP}+{\rm IDFP}+{\rm IDFN}},
\end{equation}
where ${\rm IDTP}$, ${\rm IDFP}$, and ${\rm IDFN}$ are calculated by the truth-to-result match, i.e., bipartite graph matching between true trajectories and hypothetical trajectories. Afterwards, each hypothesis is assigned to a unique target. All the frames of hypotheses with small overlap are seen as false positives and that of ground truth are seen as false negatives.

Comparing to ${\rm MOTA}$, ${\rm IDF_1}$ better measures the consistency of ID matching. A simple example to illustrate its effectiveness is presented in Figure \ref{fig2}. In this paper, we focus on the performance of identification, and thus use ${\rm IDF_1}$ as the reference of our self-evaluation metric.

\section{Self Quality Evaluation}

We design a novel self quality evaluation metric ${\rm SQE}$ to measure the tracking performance without ground truth annotations that can enable parameters optimization to gain better tracking performance in reality. This metric should be positively correlated with ${\rm IDF_1}$ which generally measures the tracking performance the best. The guiding design criteria is provided below in which we highlight some distinctive features an ideal tracker should possess. Both the theoretical and practical facets show that high-quality trajectories present single peaks in the feature distance distribution, while low-quality trajectories present multiple peaks.

\subsection{Design criteria}

To have a better understanding of the proposed metric, we first explain that an ideal MOT tracker should meet the following criteria. It should be able to: (1) track all targets continuously from appearing to leaving the tracking area; (2) track each target consistently, that is, each target should be assigned one and only one track ID over time; (3) locate the position of each target as accurately as possible. 

As mentioned in Section 2.2, (3) quantifies the detection performance in the tracking-by-detection paradigm, thus it is not our main focus. For self evaluation metrics design, (1) inspires that the number and length of trajectories are supposed to be appropriate. (2) leads to the assumption that for an outstanding tracker, REID features are as similar as possible if coming from the same trajectory, otherwise are as different as possible. This can be characterized by the intra and inter distance of trajectories. We define the distance between two features$f$ and $g$ as their Euclidean distance:
\begin{equation}\label{eq4}
distance=\lVert f-g\rVert_2.
\end{equation}

Based on the above considerations, our self evaluation metric should take the quantity, length, and feature distance information into account comprehensively. Since establishing relationship between the identification quality and the absolute values of distance is hard, distance distribution analysis is considered to be a more reasonable solution.  

\subsection{Distance distribution analysis}

We demonstrate in theory that the intra distance of the same target and the inter distance of different targets obey chi distribution. 

For object representation, it is common that low-quality inputs will lead to uncertain estimations, causing the computed REID features to fluctuate around the ideal value. We follow the assumptions in \cite{shi2019probabilistic}, modeling the distribution of features as multivariate Gaussian distribution:
\begin{equation}\label{eq5}
p(\mathbf{z})=\mathcal{N}(\mathbf{z};\bm{\mu},\bm{\sigma}^2\mathbf{I}),
\end{equation}
where $\mathbf{z}$ is a N-dimension feature vector, $\bm{\mu}$ and $\bm{\sigma}^2$ represent the ideal value and uncertainty along each dimension respectively. Each dimension obeys an independent Gaussian distribution.

We measure the Euclidean distance between a pair of features $(\mathbf{z}_i,\mathbf{z}_j)$:
\begin{equation}\label{eq6}
d(\mathbf{z}_i,\mathbf{z}_j)=\sqrt{\sum_{k=1}^N(z_{ik}-z_{jk})^2}=\sqrt{\sum_{k=1}^Nd_k^2}.
\end{equation}

According to the nature of independent Gaussian random variables, we have $p(d_k)=\mathcal{N}(d_k;\mu_{ik}-\mu_{jk},\sigma_{ik}^2+\sigma_{jk}^2)$. If $(\mathbf{z}_i,\mathbf{z}_j)$ comes from the same target, then $\mu_{ik}-\mu_{jk}=0$, $\sigma_{ik}^2+\sigma_{jk}^2=2\sigma_k^2$. Thus, the feature distance after standardization obeys chi distribution with a degree of freedom equals to N:
\begin{equation}\label{eq7}
\sqrt{\sum_{k=1}^N(\frac{d_k}{\sqrt{2\sigma_k^2}})^2}\sim\chi_N,
\end{equation}
and if $(\mathbf{z}_i,\mathbf{z}_j)$ comes from different targets:
\begin{equation}\label{eq8}
\sqrt{\sum_{k=1}^N(\frac{d_k-(\mu_{ik}-\mu_{jk})}{\sqrt{\sigma_{ik}^2+\sigma_{jk}^2}})^2}\sim\chi_N,
\end{equation}

Therefore, the intra and inter distance distributions of ideal trajectory hypotheses present single peaks. Next we consider a low-quality trajectory containing an identity switch between target A and B. For the ease of analysis, we assume that each target and feature dimension has the same variance. Therefore, the distance of features $(\mathbf{z}_{A_i},\mathbf{z}_{B_j})$ obeys non-central chi distribution with a positive noncentrality parameter $\lambda=\sum_{k=1}^N(\mu_{A_{ik}}-\mu_{B_{jk}})^2$. Meanwhile, the distance within each target obeys central chi distribution proved as above. The final distance distribution is indeed the sum of central and non-central chi distributions, thus showing a bimodal form. It can be inferred that the low-quality trajectories with wrong identification would present multiple peaks in the intra and inter distance distributions.

\subsection{Practical verification}

\begin{figure*}[h]
\centering
\subfigure[ID 0]{
    \begin{minipage}[t]{0.5\linewidth}
        \includegraphics[width=0.3\linewidth]{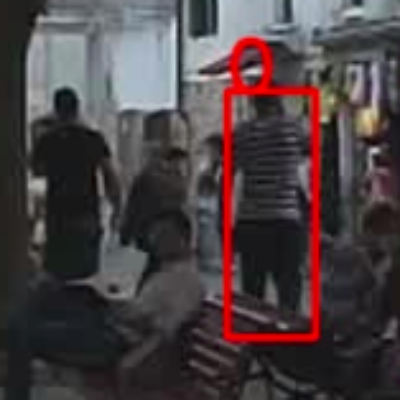}
        \vspace{0.02cm}
        \includegraphics[width=0.3\linewidth]{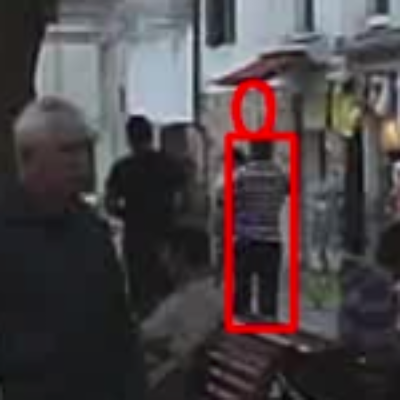}
        \vspace{0.02cm}
        \includegraphics[width=1.25in]{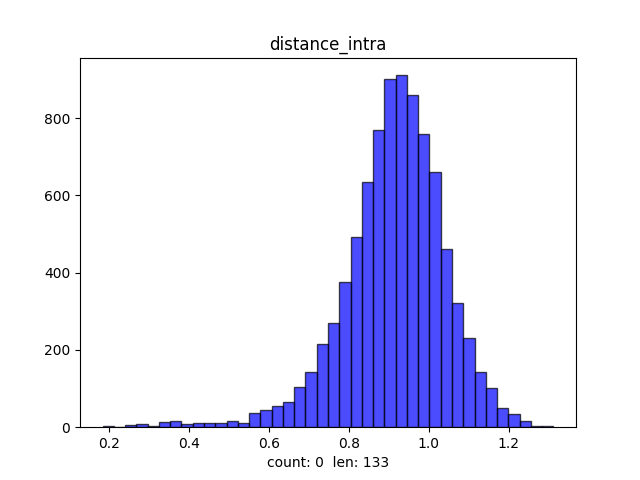}
        \vspace{0.02cm}
    \end{minipage}%
}%
\subfigure[ ID 0 \& ID 1]{
    \begin{minipage}[t]{0.5\linewidth}
        \includegraphics[width=0.3\linewidth]{figures/000_r.png}
        \vspace{0.02cm}
        \includegraphics[width=0.3\linewidth]{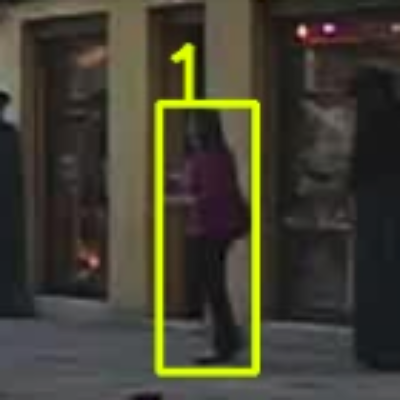}
        \vspace{0.02cm}
        \includegraphics[width=1.25in]{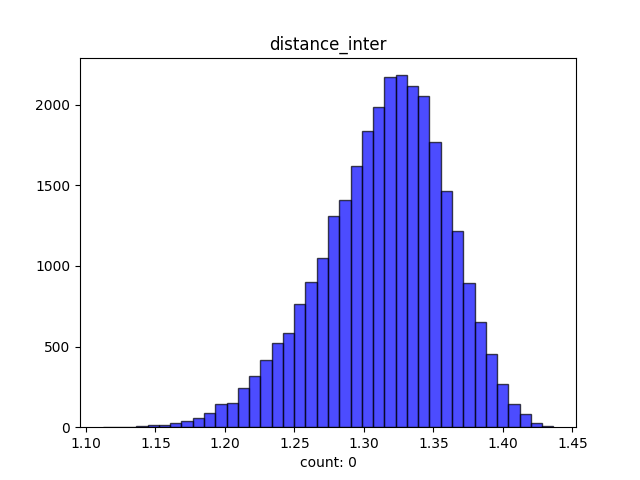}
        \vspace{0.02cm}
    \end{minipage}%
}\\%
\subfigure[ID 9]{
    \begin{minipage}[t]{0.5\linewidth}
        \includegraphics[width=0.3\linewidth]{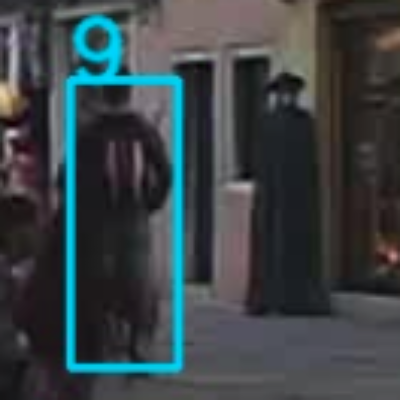}
        \vspace{0.02cm}
        \includegraphics[width=0.3\linewidth]{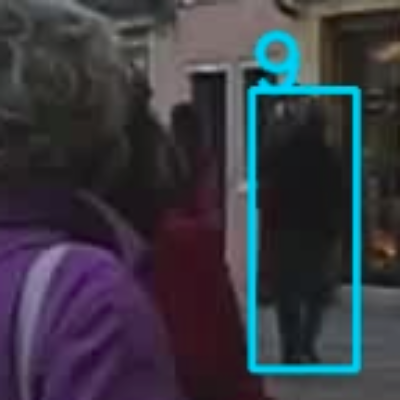}
        \vspace{0.02cm}
        \includegraphics[width=1.25in]{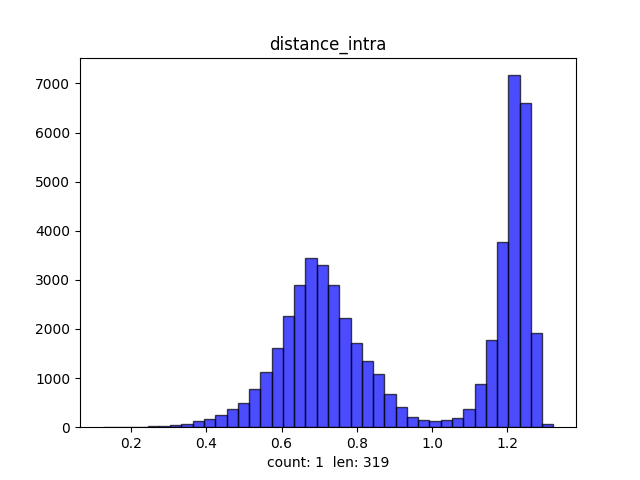}
        \vspace{0.02cm}
    \end{minipage}%
}%
\subfigure[ID 3 \& ID 220]{
    \begin{minipage}[t]{0.5\linewidth}
        \includegraphics[width=0.3\linewidth]{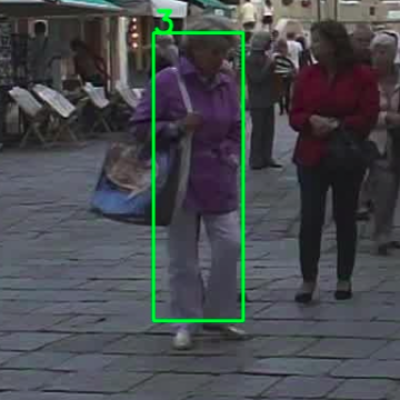}
        \vspace{0.02cm}
        \includegraphics[width=0.3\linewidth]{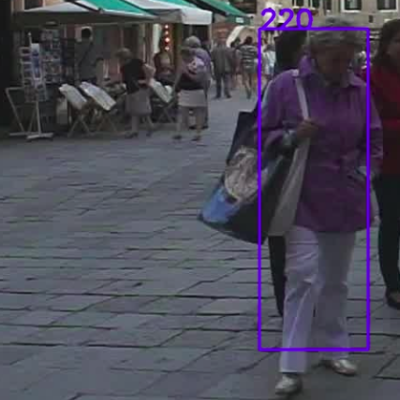}
        \vspace{0.02cm}
        \includegraphics[width=1.25in]{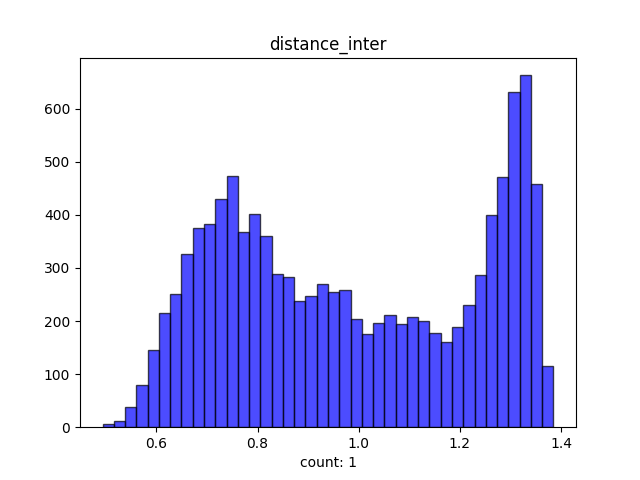}
        \vspace{0.02cm}
    \end{minipage}%
}\\%
\caption{Distance distributions of several different tracking cases and corresponding visualization results.}
\label{fig3}
\end{figure*}

We practically verify the above conclusions by visualizing the intra and inter distance distributions of several different tracking cases in Figure \ref{fig3}. The results exhibit that the high-quality trajectories, such as the one labeled with ID 0 consistently tracks a person moving forward while being separated from the one with ID 1, present single peaks. In contrast, the low-quality trajectories, such as the one containing an identity switch with ID 9 and the overlapped ones with ID 3 and ID 220, present multiple peaks.

To quantify the validity of our Gaussian assumption in Section 3.2, we use the descriptor provided in \cite{sun2018beyond} to perform a normality test on the ground truth of MOT16 train set and find that 74\% of the trajectories can be approximate as Gaussian distribution at a significance level of 0.1. Under low-density scenarios like MOT16-05, the percentage raises to 88 \%. Considering that counterexamples may occur in practice, such as two similar-dressed people, we have also tested the performance of the descriptor on classifying unique person IDs in MOT16’s detection boxes. When the precision is set to 0.95, the recall and mAP can reach 0.94 and 0.98 respectively. Therefore, we consider the counterexamples only make up a small portion.

However, due to non-ideal factors, the final distances do not fully obey the theoretical chi distribution. We take ID 0 for example. Although a similar overall shape is shown, the hypothesis test has an extremely low p-value of 0, indicating a statistically significant difference. This may have two reasons: (1) Bias is introduced when using sample statistics to replace the true mean and variance for standardization; (2) Features extracted by the REID model are not independent in each dimension. The second reason is very common, since deep neural networks tend to cause strong correlations between multiple dimensions.

It is encouraging that the trajectories of different qualities still retain the distinctive single or multiple peaks. The more frames with wrong identification, the more obvious the two peaks, and the larger interval between them. In practice we found that fitting a two-class Gaussian distribution and setting a threshold for the mean difference can qualitatively detect those low-quality trajectories which significantly affect tracking performance. According to the visualization results, we also found that the false alarm trajectory is usually short in length, large in variance, and may interfere with the inter distances to produce multiple peaks. These trajectories for which no real target exists are also categorized as low-quality trajectories.

\subsection{Metric}

\begin{algorithm}[t]
\caption{self quality evaluation}
\label{Algorithm1}
\LinesNumbered 
\KwIn{set of trajectory hypotheses $\mathcal{T}=\{T_k\}$}
\KwOut{self evaluation result}
$n$ = number of trajectories in $\mathcal{T}$\; 
$L$ = mean length of trajectories in $\mathcal{T}$\;
initialization\;
\ForEach{$T_k\in \mathcal{T}$}{
    calculate the intra distance by Equation \ref{eq4}\;
    \eIf{$L_{T_k}<\delta_L$ and $Std>\delta_D$}{
        mark the track as false alarm\;
        $fp=fp+1$\;
    }{
        fit a two-class Gaussian mixture model\;
        \If{$\Delta_{mean}>\delta_m$}{
            $dif=dif+1$\;
        }
    }
}
\ForEach{$T_l,T_m\in \mathcal{T}$}{
    calculate the inter distance by Equation \ref{eq4}\;
    \If{not false alarm tracks}{
         fit a two-class Gaussian mixture model\;
        \If{$\Delta_{mean}>\delta_m$}{
            $sim=sim+1$\;
        }
    }
}
calculate ${\rm SQE}$ by Equation \ref{eq9}\;
\textbf{return} ${\rm SQE}$\;
\end{algorithm}

Based on the above criteria and distance distribution analyses, we propose a novel self quality evaluation metric ${\rm SQE}$, which can be expressed as:
\begin{equation}\label{eq9}
{\rm SQE}=\dfrac{n\times{L}}{n+k_1\times{L}+k_2\times{(fp+dif+sim)}}
\end{equation}

The specific explanation is detailed below. The evaluation process is summarized in Algorithm \ref{Algorithm1} and mainly divided into four steps:

(1) For a trajectory with short length and large standard deviation, we mark it as false alarm and accumulate $fp$. 

(2) For the rest trajectories we utilize a two-class Gaussian mixture model to fit the intra distances, and judge whether it is a low-quality trajectory according to the mean difference. If it exceeds a certain threshold, we assert that this trajectory contains more than one target and accumulate a difference error, denoted by $dif$. 

(3) Similarly, the inter distances of each two non-false alarm trajectories are also fitted. They are considered to match the same target with a large mean difference, and the similarity error is denoted by $sim$.

(4) Other internal characteristics like the number $n$ and mean length $L$ of trajectories are also embedded.

When the REID threshold is set too strict, there are so many detection boxes being excluded that $n$ and $L$ are both small; when $n\times L$ remains almost constant, the two variables have opposite trend, and extreme situations including excessively fragmented or concatenated trajectories will lead to imbalance between them. To downgrade these poor tracking results, we employ the form of harmonic mean, and set $k_1$ to accommodate moving speed and density of tracking objects. For pedestrian tracking task on street videos, the magnitude of $n$ and $L$ is approximately equivalent, and thus $k_1$ could be set to $1$ concisely.

Based on this rough constraint form, a correction item is added to the denominator. We have demonstrated that the accumulated $dif$, $sim$ and $fp$ can reflect the number of low-quality trajectories. Therefore, their sum is expected to be small, and meanwhile the value of ${\rm SQE}$ is large. The correction item actually plays a key role within the range of moderate values of $n$ and $L$. $k_2$ is used to adjust the ratio between $n$, $L$ and sum of errors. 

Parameters in SQE are not difficult to set. $\delta_L$ is comparable to the video's frame rate. With a high-precision ReID model, randomly selecting false alarms and ID switch examples from reference videos is adequate to observe $Std$ and $\Delta_m$, so as to set $\delta_D$ and $\delta_m$ accordingly. Additionally, when the tracker and task (vehicle/pedestrian) are given, $k_1$ and $k_2$ could be set empirically. 

\section{Experiments}

{\bf Implementation details.} We assess our self evaluation method mainly on the MOT16 Challenge data sets \cite{milan2016mot16}, which contains 14 video sequences (7 for training, 7 for testing) taken by both static and moving cameras from different angles in different scenes. We focus our study on pedestrian tracking and make use of the person ReID model provided by \cite{sun2018beyond}. All the experiments are completed with the same parameter setting: $\delta_L=15$, $\delta_D=0.2$, $\delta_m=0.3$, $k_1=1$ and $k_2$ takes 2 and 10 for the REID threshold and the merging threshold, respectively. The REID threshold varies from 0.3 to 1.6, beyond which ${\rm IDF_1}$ remains invariant. Similarly, the merging threshold varies from 0.5 to 1.5. The parameter optimization process is based on grid search. The rest of this section prove the accuracy, universality, and effectiveness of our self quality evaluation metric ${\rm SQE}$.

\begin{table*}[h]

\begin{center}
\begin{threeparttable}
\setlength{\tabcolsep}{1.1mm}{
\begin{tabular}{c|ccccccc|c}

\hline
video & method & parameter & ${\rm IDF_1^*}$ & ${\rm IDP^*}$ & ${\rm IDR^*}$ & ${\rm MOTA^*}$ & ${\rm IDS^*}$ &  \makecell[c]{$\Delta$\\$({\rm IDF_1})$} \\
\hline \hline
\multirow{2}{*}{02} & baseline(gt) & 0.80 & 58.3 & 79.3 & 46.0 & 51.9 & 69 & \multirow{2}{*}{0.0} \\ \cline{2-8} & ${\rm SQE(ours)}$ & 0.80 & 58.3 & 79.3 & 46.0 & 51.9 & 69 \\ \hline
\multirow{2}{*}{04} & baseline(gt) & 1.05 & 82.0 & 93.5 & 73.0 & 77.3 & 21 & \multirow{2}{*}{1.1} \\ \cline{2-8} & ${\rm SQE(ours)}$ & 0.80 & 80.9 & 93.0 & 71.5 & 76.2 & 32 \\ \hline
\multirow{2}{*}{05} & baseline(gt) & 0.90 & 71.2 & 79.2 & 64.6 & 62.0 & 23 & \multirow{2}{*}{0.1} \\ \cline{2-8} & ${\rm SQE(ours)}$ & 1.00 & 71.1 & 78.3 & 65.2 & 61.5 & 32 \\ \hline
\multirow{2}{*}{09} & baseline(gt) & 1.20 & 76.0 & 88.8 & 66.4 & 73.5 & 8 & \multirow{2}{*}{1.3} \\ \cline{2-8} & ${\rm SQE(ours)}$ & 0.80 & 74.7 & 88.6 & 64.6 & 72.2 & 7 \\ \hline
\multirow{2}{*}{10} & baseline(gt) & 0.95 & 72.4 & 76.6 & 68.7 & 71.5 & 79 & \multirow{2}{*}{1.9} \\ \cline{2-8} & ${\rm SQE(ours)}$ & 0.90 & 70.5 & 74.5 & 66.8 & 71.2 & 81 \\ \hline
\multirow{2}{*}{11} & baseline(gt) & 0.85 & 80.1 & 89.7 & 72.4 & 75.0 & 29 & \multirow{2}{*}{4.6} \\ \cline{2-8} & ${\rm SQE(ours)}$ & 1.00 & 75.5 & 83.8 & 68.7 & 73.5 & 34 \\ \hline
\multirow{2}{*}{13} & baseline(gt) & 0.75 & 58.2 & 74.6 & 47.7 & 47.0 & 73 & \multirow{2}{*}{2.3} \\ \cline{2-8} & ${\rm SQE(ours)}$ & 1.05 & 55.9 & 68.9 & 47.0 & 45.6 & 90 \\ \hline

\end{tabular}}
\begin{tablenotes}
    \footnotesize
    \item[*] denotes that the score for ${\rm SQE}$ parameters is only calculated after the parameters are determined by ${\rm SQE}$, but not used to tune the parameters.
\end{tablenotes}
\end{threeparttable}
\end{center}
\caption{Comparison of the optimal REID threshold determined by ground truth and our self evaluation method on MOT16-02$\sim$13.}
\label{table1}
\end{table*}

\begin{figure}[h]
\centering
\subfigure[]{
    \begin{minipage}[t]{0.5\linewidth}
        \centering
        \includegraphics[width=1.6in]{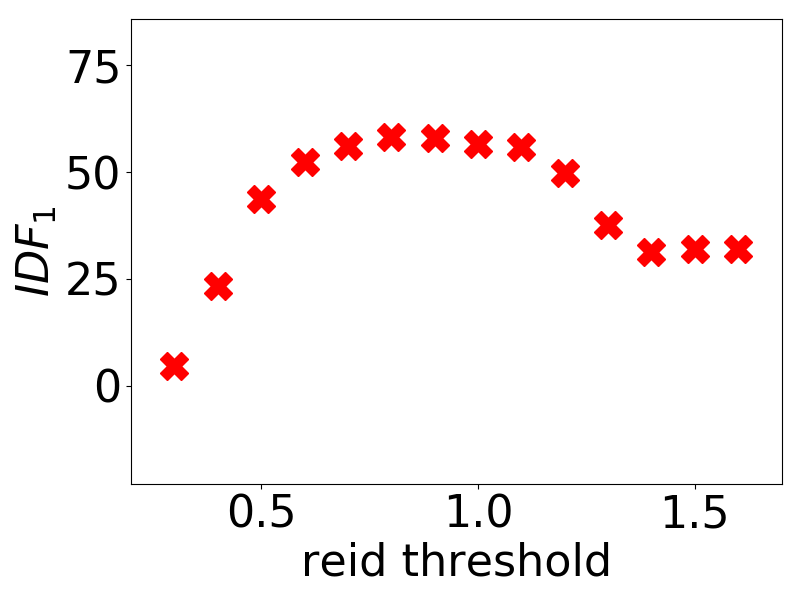}\\
        \vspace{0.02cm}
    \end{minipage}%
}%
\subfigure[]{
    \begin{minipage}[t]{0.5\linewidth}
        \centering
        \includegraphics[width=1.6in]{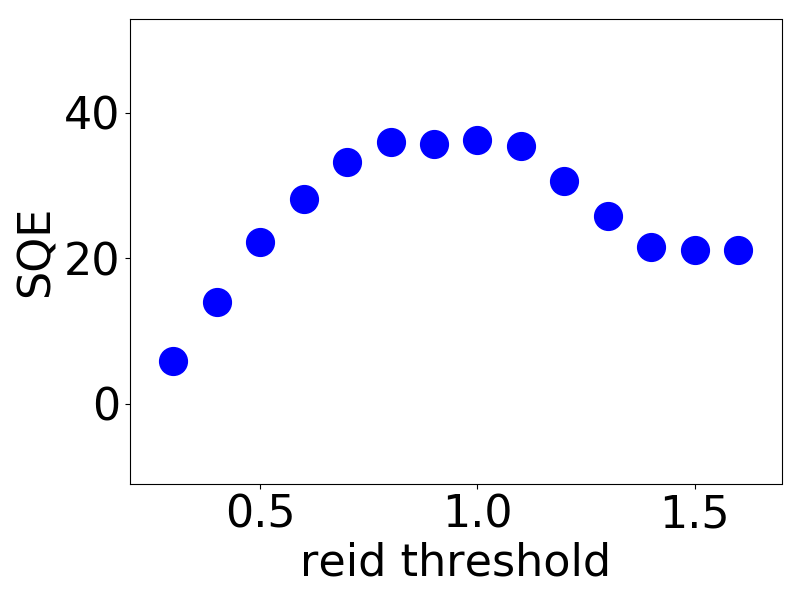}\\
        \vspace{0.02cm}
    \end{minipage}%
}%
\centering
\caption{Visualization of ${\rm IDF_1}$ and ${\rm SQE}$ on MOT16-02 (complex scene) when changing the REID threshold.}
\label{fig5}
\end{figure}

{\bf Comparison with supervised metrics.} To demonstrate the effectiveness of our self evaluation metric in evaluating tracking performance, we compare its score with existing commonly used supervised metrics on MOT16-02 training video, and visualize ${\rm IDF_1}$ and ${\rm SQE}$ in Figure \ref{fig5}. We found that as the the REID threshold ascends, both ${\rm SQE}$ and ${\rm IDF_1}$ increase at first and decrease afterwards, and reach the highest value at 0.8 with relatively high ${\rm IDP}$, ${\rm IDR}$, and ${\rm MOTA}$. These two items present a very similar trend, which indicates that our designed metric can primarily fulfill the desired positive correlations with ${\rm IDF_1}$ which generally measures the performance of identification the best.

\begin{figure}[h]
\centering
\subfigure[]{
    \begin{minipage}[t]{0.5\linewidth}
        \centering
        \includegraphics[width=1.6in]{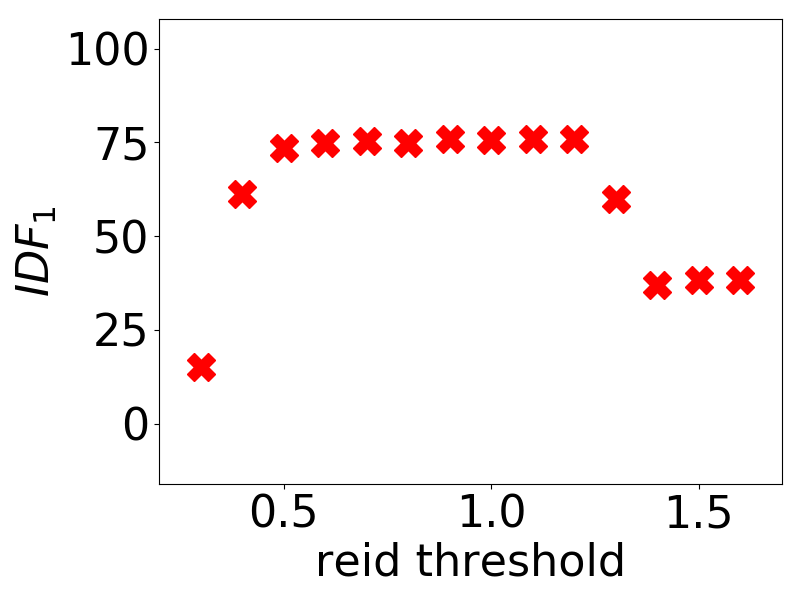}\\
        \vspace{0.02cm}
    \end{minipage}%
}%
\subfigure[]{
    \begin{minipage}[t]{0.5\linewidth}
        \centering
        \includegraphics[width=1.6in]{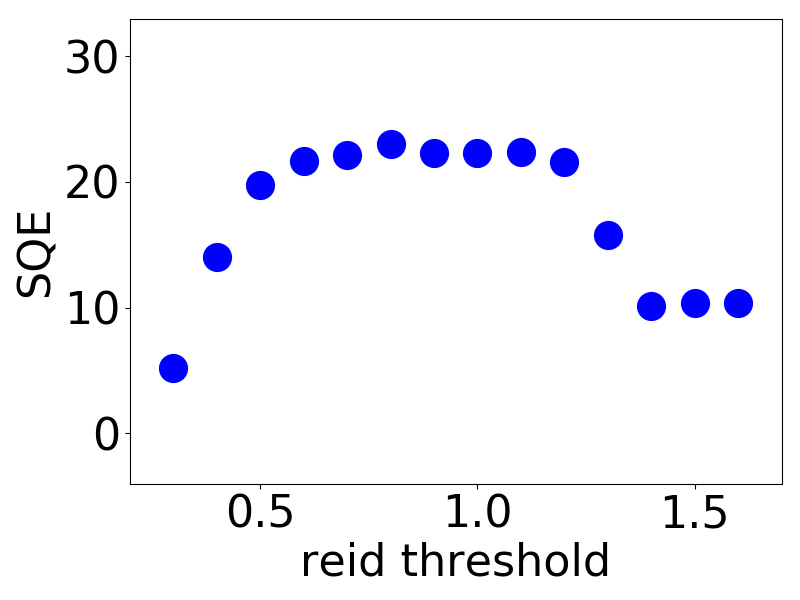}\\
        \vspace{0.02cm}
    \end{minipage}%
}%
\centering
\caption{Visualization of ${\rm IDF_1}$ and ${\rm SQE}$ on MOT16-09 (simple scene) when changing the REID threshold.}
\label{fig6}
\end{figure}

\begin{figure}[h]
\centering
\subfigure[]{
    \begin{minipage}[t]{0.5\linewidth}
        \centering
        \includegraphics[width=1.6in]{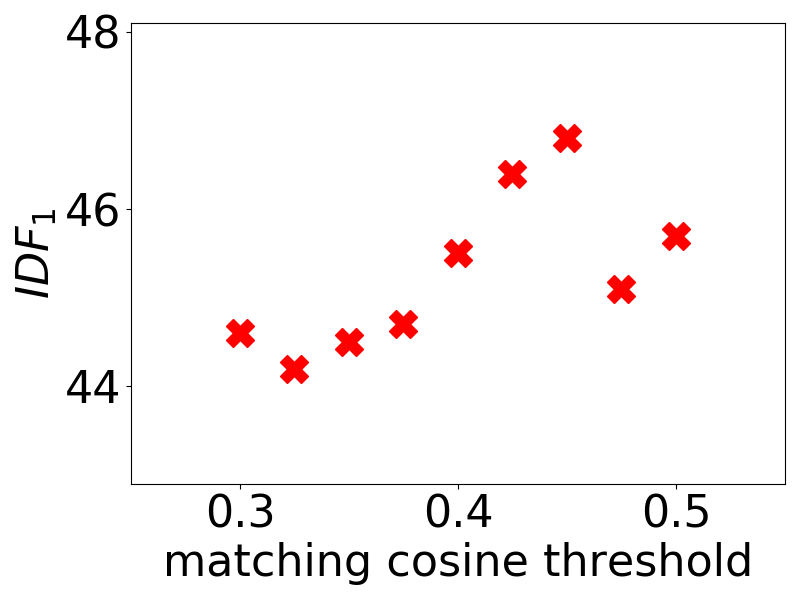}\\
        \vspace{0.02cm}
    \end{minipage}%
}%
\subfigure[]{
    \begin{minipage}[t]{0.5\linewidth}
        \centering
        \includegraphics[width=1.6in]{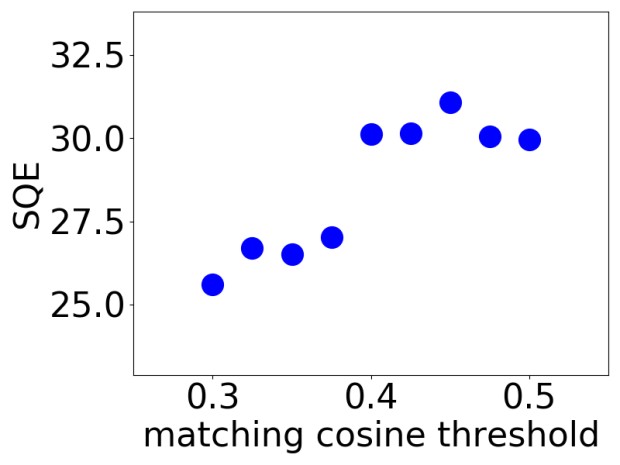}\\
        \vspace{0.02cm}
    \end{minipage}%
}%
\centering
\caption{Visualization of ${\rm IDF_1}$ and ${\rm SQE}$ on MOT16-02 when changing the matching cosine threshold in Deep SORT algorithm.}
\label{fig7}
\end{figure}

MOT16-02 video records a complex scene with a large number of people walking around a large square. We further analyse the result on MOT16-09 video, a simpler street scene with low density and the least number of tracks from a low angle, in Figure \ref{fig6}. The favourable similarity illustrates that our self evaluation method can be generalized to different viewpoints and scenarios. The detailed results on other videos are provided in the supplementary material. We summarize the optimal REID threshold determined by ${\rm IDF_1}$ and ${\rm SQE}$ in Table \ref{table1}, with corresponding evaluation scores under these parameters. Our self evaluation method can approximately quantify tracking performance, specifically, 85\% of the optimal parameter differences do not exceed 0.25, and 85\% of the corresponding ${\rm IDF_1}$ differences do not exceed 3.

{\bf Generalization to other tracking algorithms.} To illustrate the robustness and universality of our method, other tracking algorithms are supposed to be tested as a supplementary experiment. We choose Deep SORT, which is one of the highly recognized and open source MOT algorithms in recent years. The REID threshold corresponds to the matching cosine threshold in Deep SORT. This algorithm replaces our interpolation logic with IOU matching, causing the features during occlusion time period to exhibit a small interference peak in the intra distance distribution; therefore, we remove the feature information of these frames when performing self evaluation. As shown in Figure \ref{fig7}, a strong correlation between ${\rm IDF_1}$ and ${\rm SQE}$ is presented, demonstrating the success of our method on other trackers.

\begin{figure}[]
\centering
\subfigure[]{
    \begin{minipage}[t]{0.5\linewidth}
        \centering
        \includegraphics[width=1.6in]{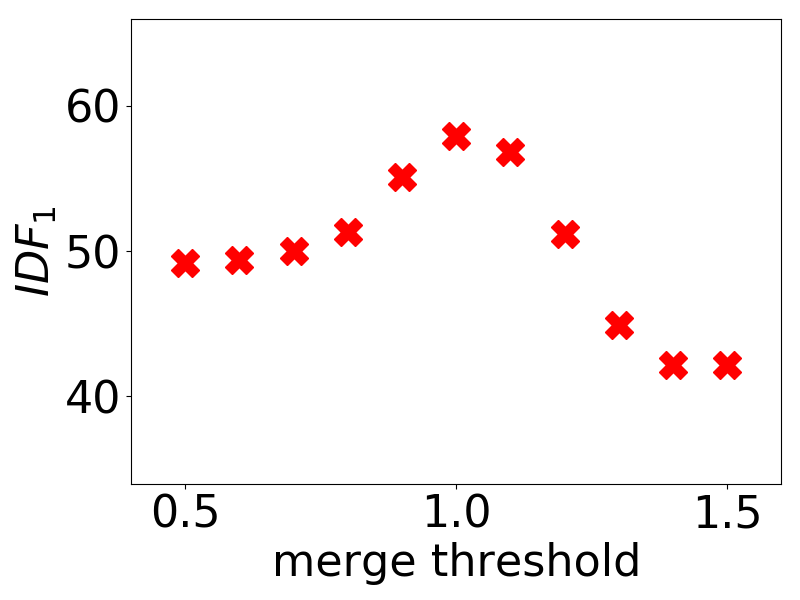}\\
        \vspace{0.02cm}
    \end{minipage}%
}%
\subfigure[]{
    \begin{minipage}[t]{0.5\linewidth}
        \centering
        \includegraphics[width=1.6in]{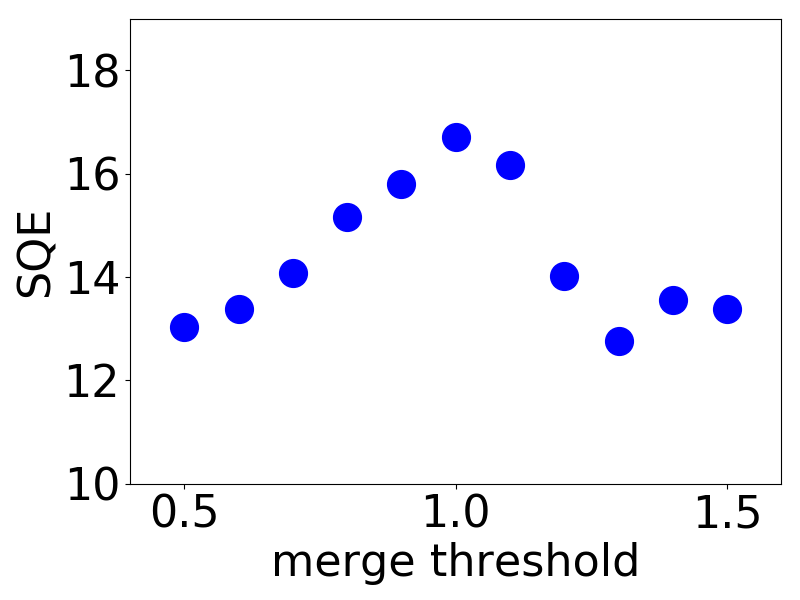}\\
        \vspace{0.02cm}
    \end{minipage}%
}%
\centering
\caption{Visualization of ${\rm IDF_1}$ and ${\rm SQE}$ on MOT16-02 (complex scene) when changing the merging threshold.}
\label{fig8}
\end{figure}

\begin{figure}[]
\centering
\subfigure[]{
    \begin{minipage}[t]{0.5\linewidth}
        \centering
        \includegraphics[width=1.6in]{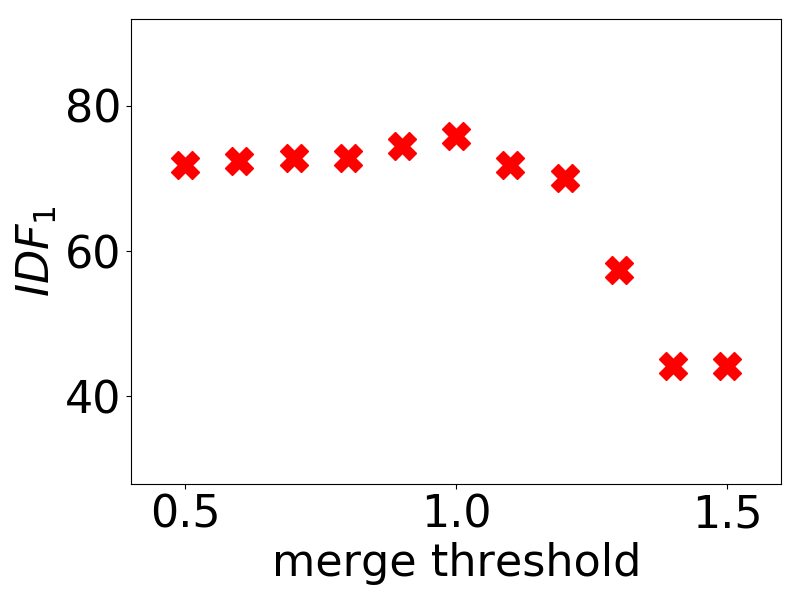}\\
        \vspace{0.02cm}
    \end{minipage}%
}%
\subfigure[]{
    \begin{minipage}[t]{0.5\linewidth}
        \centering
        \includegraphics[width=1.6in]{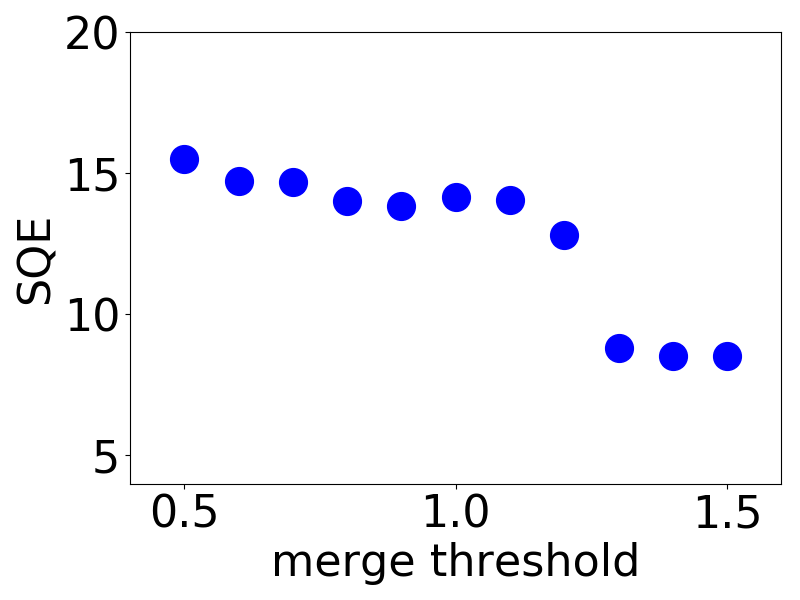}\\
        \vspace{0.02cm}
    \end{minipage}%
}%
\centering
\caption{Visualization of ${\rm IDF_1}$ and ${\rm SQE}$ on MOT16-09 (simple scene) when changing the merging threshold.}
\label{fig9}
\end{figure}

{\bf Generalization to other parameters.} We further test the universality of our method on other parameters. Except for the REID threshold, the merging threshold is another dominant factor affecting final tracking performance. Similarly, we visualize the comparison of ${\rm IDF_1}$ and ${\rm SQE}$ of both complex and simple scenes in Figure \ref{fig8} and \ref{fig9}. The results still maintain positive correlations. Table \ref{table2} shows a high accuracy, with 5 out of 7 videos have an optimal parameter difference below 0.1, and almost all the corresponding ${\rm IDF_1}$ differences do not exceed 3.

\begin{table*}[t]

\begin{center}
\setlength{\tabcolsep}{1.1mm}{
\begin{tabular}{c|ccccccc|c}

\hline
video & method & parameter & ${\rm IDF_1^*}$ & ${\rm IDP^*}$ & ${\rm IDR^*}$ & ${\rm MOTA^*}$ & ${\rm IDS^*}$ &  \makecell[c]{$\Delta$\\$({\rm IDF_1})$} \\
\hline \hline
\multirow{2}{*}{02} & baseline(gt) & 1.00 & 57.9 & 72.7 & 46.1 & 51.7 & 82 & \multirow{2}{*}{0.0} \\ \cline{2-8} & ${\rm SQE(ours)}$ & 1.00 & 57.9 & 72.7 & 46.1 & 51.7 & 82 \\ \hline
\multirow{2}{*}{04} & baseline(gt) & 1.05 & 82.3 & 94.2 & 73.1 & 76.8 & 23 & \multirow{2}{*}{0.6} \\ \cline{2-8} & ${\rm SQE(ours)}$ & 1.00 & 81.7 & 93.7 & 72.4 & 76.5 & 27 \\ \hline
\multirow{2}{*}{05} & baseline(gt) & 0.85 & 73.6 & 82.5 & 66.5 & 61.9 & 34 & \multirow{2}{*}{0.5} \\ \cline{2-8} & ${\rm SQE(ours)}$ & 0.75 & 73.1 & 82.8 & 65.5 & 61.4 & 46 \\ \hline
\multirow{2}{*}{09} & baseline(gt) & 1.00 & 75.9 & 89.2 & 66.0 & 73.2 & 7 & \multirow{2}{*}{3.1} \\ \cline{2-8} & ${\rm SQE(ours)}$ & 0.70 & 72.8 & 85.8 & 63.2 & 72.8 & 12 \\ \hline
\multirow{2}{*}{10} & baseline(gt) & 1.05 & 71.5 & 75.1 & 68.2 & 70.8 & 77 & \multirow{2}{*}{1.6} \\ \cline{2-8} & ${\rm SQE(ours)}$ & 0.95 & 69.9 & 74.5 & 65.9 & 71.5 & 82 \\ \hline
\multirow{2}{*}{11} & baseline(gt) & 1.10 & 78.2 & 87.6 & 70.5 & 75.0 & 30 & \multirow{2}{*}{2.8} \\ \cline{2-8} & ${\rm SQE(ours)}$ & 0.85 & 75.4 & 84.7 & 68.0 & 73.9 & 35 \\ \hline
\multirow{2}{*}{13} & baseline(gt) & 1.05 & 56.7 & 70.7 & 47.3 & 46.2 & 68 & \multirow{2}{*}{0.6} \\ \cline{2-8} & ${\rm SQE(ours)}$ & 1.10 & 56.1 & 69.2 & 47.1 & 45.4 & 72 \\ \hline

\end{tabular}}
\end{center}
\caption{Comparison of the optimal merging threshold determined by ground truth and our self evaluation method on MOT16-02$\sim$13.}
\label{table2}
\end{table*}


{\bf Practical testing.} Our ultimate goal is to find the optimal parameters in realistic scenes where ground truth is unavailable. Additionally in reality the training data is relatively small in scale comparing to the unknown test environment. To test our method in a pragmatic manner, we regard the first 4 training videos as our test set and the last 3 training videos as our training set. Conventionally, the parameters are tuned on the training set and remain constant during testing. In our simulation, we name these parameters as the baseline parameters. Conversely, our ${\rm SQE}$ metric can guide the self-optimization of parameters without ground truth. Thus, it is employed directly to tune the 4 testing videos individually. 

In reality we can first acquire baseline parameters as reference on small-scale training data, then conduct self evaluation to further optimize the parameters in a relatively small range. The procedure of computing the customized parameters is as follows: (1) Find baseline parameters; (2) For each testing video, fix one parameter with reference to baseline, and then tune the other according to ${\rm SQE}$ alternately; (3) Combine them to be the customized parameters. 

Our method is considered to be effectual if the tracker using the customized parameters outperform the tracker using constant training-set-tuned parameters. The result is shown in Table \ref{table3}, where gt denotes the true optimal parameters on each video. To be rigorous, we use the best parameters found by grid search on the 3 assumed training videos as the baseline. It is apparent that the parameters tuned by ${\rm SQE}$ achieve considerable improvement comparing to the baseline, and the results are much closer to the the true optimum, showing the effectiveness of our method when implemented in a practical manner.

\begin{table*}[t]
\begin{center}
\begin{tabular}{c|ccccccc}
\hline
video & method & parameter & ${\rm IDF_1^*}$ & ${\rm IDP^*}$ & ${\rm IDR^*}$ & ${\rm MOTA^*}$ & ${\rm IDS^*}$ \\
\hline \hline
\multirow{3}{*}{02} & gt & 0.85, 0.95 & 59.2 & 79.9 & 47.0 & 52.8 & 80 \\ \cline{2-8} & baseline & 0.75, 1.00 & 56.2 & {\bf 78.3} & 43.8 & 50.1 & {\bf 61} \\ \cline{2-8} & ${\rm SQE(ours)}$ & 0.90, 1.00 & {\bf 57.9} & 75.5 & {\bf 47.0} & {\bf 52.8} & 82 \\ \hline
\multirow{3}{*}{04} & gt & 0.85, 1.05 & 82.6 & 94.6 & 73.3 & 76.7 & 23 \\ \cline{2-8} & baseline & 0.75, 1.05 & 80.9 & 92.9 & 71.6 & 76.3 & 25 \\ \cline{2-8} & ${\rm SQE(ours)}$ & 0.75, 1.10 & {\bf 82.3} & {\bf 94.4} & {\bf 72.9} & {\bf 76.6} & {\bf 21} \\ \hline
\multirow{3}{*}{05} & gt & 0.90, 0.85 & 73.6 & 82.5 & 66.5 & 61.9 & 34 \\ \cline{2-8} & baseline & 0.75, 1.05 & 68.4 & 80.1 & 59.7 & 57.6 & {\bf 24} \\ \cline{2-8} & ${\rm SQE(ours)}$ & 1.00, 0.95 & {\bf 72.2} & {\bf 81.1} & {\bf 65.1} & {\bf 62.6} & 25 \\ \hline
\multirow{3}{*}{09} & gt & 1.20, 0.70 & 76.0 & 88.8 & 66.4 & 73.5 & 8 \\ \cline{2-8} & baseline & 0.75, 1.05 & 71.4 & 84.7 & 61.7 & {\bf 72.2} & {\bf 6} \\ \cline{2-8} & ${\rm SQE(ours)}$ & 0.85, 0.90 & {\bf 73.0} & {\bf 87.1} & {\bf 62.8} & 71.2 & 10 \\  \hline
\multirow{3}{*}{overall} & gt & - & 76.4 & 90.3 & 66.2 & 69.7 & 145 \\ \cline{2-8} & baseline & - & 74.0 & 88.5 & 63.5 & 68.4 & {\bf 116} \\ \cline{2-8} & ${\rm SQE(ours)}$ & - & {\bf 75.5} & {\bf 88.8} & {\bf 65.6} & {\bf 69.5} & 138 \\ \hline
\end{tabular}
\end{center}
\caption{Comparison of different REID and merging threshold selections based on ground truth, optimum of the remaining 3 videos and our ${\rm SQE}$ metric. Adjust two parameters at the same time.}
\label{table3}
\end{table*}

\begin{table}[t]
\begin{center}
\begin{threeparttable}
\begin{tabular}{c|ccccc}
\hline
dataset & method & ${\rm IDF_1^*}$ & ${\rm IDP^*}$ & ${\rm IDR^*}$ & ${\rm IDS^*}$ \\
\hline\hline
\multirow{2}{*}{MOT16 test} & baseline & 66.6 & 75.8 & 59.4 & 442 \\ \cline{2-6} & ours & {\bf 68.3} & 83.4 & 57.8 & 456 \\ \hline
\multirow{2}{*}{KITTI train$^{**}$} & baseline & 67.4 & 67.2 & 67.7 & 37 \\ \cline{2-6} & ours & {\bf 68.5} & 67.9 & 69.1 & 44 \\ \hline
\end{tabular}
\begin{tablenotes}
        \footnotesize
        \item[**] we use the 5 videos with the most pedestrians in KITTI train set.
\end{tablenotes}
\end{threeparttable}
\end{center}
\caption{Practical testing on MOT16 test set and KITTI train set.}
\label{table4}
\end{table}

To further illustrate the performance of self-optimization using ${\rm SQE}$, we experiment on MOT16 test set and KITTI train set \cite{kitti}.  The baseline parameters are the best parameters found by grid search on MOT16 training set, which outperform empirical parameters by 5.8\% ${\rm IDF_1}$ already. This setup is based on the updated  submitting policy of KITTI \footnote{http://www.cvlibs.net/datasets/kitti/eval\_tracking.php}, and we believe it can simulate pedestrian tracking in reality where test scenes varies greatly compared to annotated videos. As shown in Table \ref{table4}, the parameter self-optimization enabled by ${\rm SQE}$ elevates the performance of the tracker on these data sets.

{\bf Drawbacks and prospects.} The above experiments reflect the effectiveness of our proposed ${\rm SQE}$ metric, while there are still some drawbacks worth noting. Firstly, due to the randomness during model fitting, $dif$ and $sim$ possess several units of uncertainty, resulting in insufficient sensitivity to small changes in ${\rm IDF_1}$. Secondly, current metrics lack physical consistency explanation. ${\rm IDF_1}$ is calculated by ${\rm IDTP}$, ${\rm IDFP}$ and ${\rm IDFN}$, while our method simply records the number of low-quality trajectories. A more precise idea is to estimate ${\rm IDTP}$ and ${\rm IDFP}$ relying on the quantity information. Assume that for a trajectory where an identity switch occurs, target A appears $n_1$ frames while target B appears $n_2$ frames. The total length is $L$ and the number of distances in the class with larger values is $N$. Then A and B satisfy the following conditions:
\begin{equation}
\begin{cases}
n_1+n_2=L,\\
n_1\times n_2=N,
\end{cases}
\end{equation}
which can be easily solved. We can make estimations by:
\begin{equation}
{\rm IDTP},{\rm IDFP}=\max{(n_1,n_2)},\min{(n_1,n_2)}.
\end{equation}

Such processing for the intra distance distribution can accurately estimate the number of erroneous frames. Furthermore, the inter distance distribution can help refine the estimations. For example, if there is another trajectory that also tracks A, we only keep the longer one as ${\rm IDTP}$ according to the calculation rule of ${\rm IDF_1}$. However, more detailed considerations are needed for global precise estimations. In addition, categorizing low-quality trajectories and estimating erroneous frames may also be conducive to tracker's post-processing so as to improve tracking performance. Finally, the adjustable parameters $k_1$ and $k_2$ need to be defined more strictly. We plan to investigate these downsides in the future.

\section{Conclusion}

In this paper, we propose a self quality evaluation metric ${\rm SQE}$ to enable the parameters optimization in the test environment and realistic scenes where ground truth is unavailable. This new perspective can bypass the difficulty of designing an algorithm that perform well in various scenes. We demonstrate that trajectories with different qualities exhibit different single or multiple peaks in feature distance distribution, inspiring us to use a two-class Gaussian mixture model to estimate identification errors. Experiments mainly on the MOT16 Challenge data sets demonstrate the effectiveness of our method in both correlating with existing metrics and enabling parameters self-optimization to achieve better tracking performance. In the end, the drawbacks and prospects for future work are summed up. We believe that our work is instructive for further MOT research. 

\section{Acknowledgement}

This research was supported by National Key R\&D Program of China (No. 2017YFA0700800).

{\small
\bibliographystyle{ieee_fullname}
\bibliography{egbib}
}

\end{document}